\documentclass[conference]{IEEEtran}
\IEEEoverridecommandlockouts
\usepackage{cite}
\usepackage{amsmath,amssymb,amsfonts}
\usepackage[linesnumbered,ruled]{algorithm2e}
\usepackage{algorithmic}
\usepackage{graphicx}
\usepackage{textcomp}
\usepackage{xcolor}
\usepackage{amsmath}
\usepackage{comment}
\usepackage{booktabs} 
\usepackage{nicefrac}
\usepackage{textcomp}
\usepackage{booktabs}
\usepackage{caption}
\usepackage{subcaption}
\usepackage{float}

  \usepackage{enumitem}  
    \newlist{inlinelist}{enumerate*}{1}
\setlist*[inlinelist,1]{%
  label=\arabic*),
}

    \newlist{inlinelistroman}{enumerate*}{1}
\setlist*[inlinelistroman,1]{%
  label=\roman*),
}

\colorlet{annotated}{black}

\newlist{inlineitemlist}{enumerate*}{1}
\setlist*[inlineitemlist,1]{%
  label=\null,
}
    
\usepackage{nomencl}
\makenomenclature

\makeatletter
\newcommand*\dashline{\rotatebox[origin=c]{90}{$\dabar@\dabar@\dabar@$}}

\usepackage{hyperref}
\newcommand{\orcid}[1]{\href{https://orcid.org/#1}{\includegraphics[width=8pt]{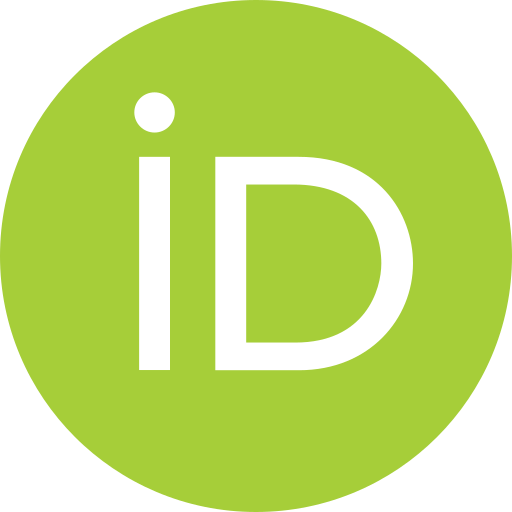}}}

\newcommand{\beginsupplement}{%
        \setcounter{table}{0}
        \renewcommand{\thetable}{S\arabic{table}}%
        \setcounter{figure}{0}
        \renewcommand{\thefigure}{S\arabic{figure}}%
     }

\begin{document}

\title{Fast mesh denoising with data driven normal filtering using deep variational autoencoders 
}


\author{
\IEEEauthorblockN{Stavros Nousias\orcid{0000-0002-2811-235X}, \textit{Member, IEEE}, and Gerasimos Arvanitis\orcid{0000-0001-8149-5188}, and Aris S. Lalos\orcid{0000-0003-0511-9302}, \textit{Senior Member, IEEE},\\ and Konstantinos Moustakas\orcid{0000-0001-7617-227X}, \textit{Senior Member, IEEE}
} 
\thanks{This work was supported by European Union Horizon 2020 Research and innovation program "WARMEST -or loW Altitude Remote sensing for the Monitoring of the state of cultural hEritage Sites: building an inTegrated model for maintenance" under Marie Sklodowska grant agreement No 777981.}
\thanks{This work was supported by European Union Horizon 2020 Research and innovation program Ageing@Work under Ageing@Work  grant agreement No 826299.}

\thanks{Stavros Nousias is with the Industrial Systems Institute, Athena Research Center, Stadiou Platani Rion Patras, 26504, Greece and the Department of Electrical \& Computer Engineering University of Patras, Rion Patras, 26504, Greece (e-mail: nousias@isi.gr, nousias@ece.upatras.gr)}
\thanks{Aris Lalos is with the Industrial Systems Institute, Athena Research Center, Stadiou Platani Rion Patras, 26504, Greece (e-mail: lalos@isi.gr)}
\thanks{Gerasimos Arvanitis and Konstantinos Moustakas are with the Department of Electrical \& Computer Engineering University of Patras, Rion Patras, 26504, Greece (e-mail:\{arvanitis,
moustakas\}@ece.upatras.gr)}
}


\maketitle

\begin{abstract} Recent advances in 3D scanning technology have enabled the deployment of 3D models in various industrial applications like digital twins, remote inspection and reverse engineering. Despite their evolving performance, 3D scanners, still introduce noise and artifacts in the acquired dense models. In this work, we propose a fast and robust denoising method for dense 3D scanned industrial models. The proposed approach employs conditional variational autoencoders to effectively filter face normals. Training and inference are performed in a sliding patch setup reducing the size of the required training data and execution times. We conducted extensive evaluation studies using 3D scanned and CAD models. The results verify plausible denoising outcomes, demonstrating similar or higher reconstruction accuracy, compared to other state-of-the-art approaches. Specifically, for 3D models with more than $1e4$ faces, the presented pipeline is twice as fast as methods with equivalent reconstruction error.
\end{abstract}

\begin{IEEEkeywords}
3D mesh denoising, data driven normal filtering, variational autoencoders. 
\end{IEEEkeywords}

\section{Introduction}


Industrial sites and manufacturing plants, often require infrastructure upgrades and construction projects. Modifications in the production line lead to downtime, high costs or unwanted delays. Digital twins could allow for improved supervision, inspection and monitoring based on simulation studies. Subsequently, they could enable the identification of errors while handling risk and dealing with liabilities. 3D scanning can facilitate the development of accurate, high-end digital twins of manufacturing processes and factory layouts.

\color{annotated}
Furthermore, smart manufacturing encompasses "fully-integrated, collaborative manufacturing systems that respond in real-time to meet changing demands and conditions in the factory, in the supply network, and in customer needs" \cite{lu2016current,nist2014smart}.
\color{black}
Accurate digital reconstruction for material inspection, quality control and reverse engineering are challenges in this evolving landscape. Inline 3D scanning could allow the examination of constructed parts, in many stages of the manufacturing process. 
\cite{agrawal2018guest,li2018deep,iqbal2019fault,wang2018deepsmart}.

Towards this direction, technologies in metrology have changed, in the past few decades, from stand-alone coordinate measuring machines (CCMs) to portable 3D scanners. The benefits of 3D scanning become evident in several use cases. Quality assurance protocols in automotive industry\cite{li2018deep,von2016multiresolution}, maintenance processes in maritime industry\cite{fraga2018review} and automated reverse engineering prove that error-free representations are a requirement for Industry 4.0 outcomes.

Several use cases appear in the literature, i.e. Artec 3D reports that a foundry\cite{artec3d2019howonefountry} uses handheld solutions to scan 3D castings, saving time and increasing productivity. The authors in \cite{aanaes2018autonomous} present a use case of a surface inspection method for wind turbines, employing an autonomous robotic arm equipped with a 3D scanner.
Moreover, state-of-the-art 3D industrial scanning outcomes, available online \cite{artec3d2019industrial} reveal that models of 2 million sampled points require up to $30$ minutes for scanning and up to $30$ minutes for post-processing. Robust, high accuracy and low-error processing outcomes would reduce the scanning time since the collection of fewer samples would be necessary for equivalent results, while fast processing would reduce the post-processing times.

These challenging issues highlight the need for parallelizable computationally inexpensive, and accurate approaches for mesh denoising. 
In a classic scenario, scanners yield noisy point clouds that are consequently converted to noisy 3D meshes. 
Denoising aims to remove the noise while preserving features and multi-scale geometric details. 
Noise is usually inserted by scanning devices and digitization processes, thus making mesh denoising an important post-processing step.
Several methods are available in the literature with significant denoising results\cite{zheng2011bilateral,zhang2015guided,sun2007fast,he2013mesh,wang2016mesh}. 
Yet the need for robust and fast algorithms, able to handle dense models rapidly, becomes essential in industrial applications\cite{bosche2010automated,srinivasan2015automatic}, where they are expected to significantly reduce the operational cost of many manufacturing tasks.

\color{annotated}
Motivated by the aforementioned challenges, we provide a fast approach for mesh denoising, based on data-driven normal filtering. We employ deep conditional variational autoencoders allowing to handle efficiently dense models. We summarize the contributions of the proposed approach in the following points:
\begin{itemize}
    \item The network can localize since, training and inference are performed in a sliding patch setup. The filtered face normal vector is generated by providing a patch of neighbouring faces as input, corresponding to a local region around that face.
    \item It requires a relatively small training set. We propose a preprocessing method that describes each patch with a scale, translation and rotation invariant representation.
    \item Evaluation studies indicate that our approach demonstrates lower complexity and execution times than other non-data-driven state-of-the-art methods. Specifically, for 3D models with more than $10^4$ faces, the presented pipeline is twice as fast as methods with an equivalent reconstruction error.
    \item It is fully parallelizable. We evaluated the execution efficiency with respect to the number of utilized cores and type of processing unit (i.e., CPU or GPU). 
    \item It can be employed for feature-preserving denoising of dense 3D models, with different noise patterns. Such a property would be ideally suited for industrial applications where 3D scanners with different properties are generating dense representations of physical objects.
    \item It is parameter-free since every used parameter is predefined and the user does not need to search for optimal values per model.
\end{itemize}
\color{black}
Evaluation studies were carried out using scanned and CAD 3D industrial models. Our results verify the effectiveness of the proposed method, compared to other state-of-the-art approaches, both in terms of denoising quality and computational efficiency.

The rest of this paper is organized as follows: Section II presents state-of-the-art methods and related works. Section III focused on preliminaries. Section IV describes the workflow of the proposed approach in detail. Section V is dedicated to the experimental setup and simulation results, while conclusions are drawn in Section VI.

\section{Related Work}
Mesh denoising approaches can be organized in the following categories: isotropic and anisotropic mesh filtering based, regularization based and data-driven methods.
 \subsubsection*{Isotropic and anisotropic mesh filtering}
Laplacian 
and Taubin 
smoothing are well-known approaches that remove noise and artifacts by employing iterative vertex update based on the Laplacian matrix of the geometry. To the same direction, Desbrun et al. used the mean curvature flow
\cite{desbrun1999implicit} to allow treating of irregular surfaces. However, one of their main disadvantages is that they do not preserve geometric features.  
Another category of methods, namely graph spectral processing \cite{ zhang2010spectral}, employs singular values, eigenvectors and eigenspace projections to separate 3D mesh data from noise. 
However, their disadvantage is that they are computationally expensive and resource consuming. 
Other widely accepted feature preserving approaches \cite{wang2015rolling} process vertex positions locally while preserving the geometric features. The main drawback is, in many cases, the deformation of large scale features. 
Mesh bilateral filtering methods use normal coordinates to estimate the parameters of noise removing filters, with the reconstruction accuracy relying heavily on noise characteristics \cite{fleishman2003bilateral}.
Such approaches are based on normal filtering and vertex position update \cite{zhang2015guided, sun2007fast, zheng2011bilateral, arvanitis2019feature} consisting of two iterative stages. The first stage filters the face normals while the second updates the position of vertices. 
Although this category of approaches preserves most of the sharp features, they require heavy parameterization and fail to generalize. 

\noindent \subsubsection*{Regularization based}
Regularizers are often used for ill-posed problems. Denoising of 3D meshes is in many cases an ill-posed problem due to sensing limitations and non-uniform sampling operations.
Zhang et al. \cite{zhang2015variational} minimize the energy of both vertex position and normal error and He et al.\cite{he2013mesh}, propose an $L_0$ minimization approach. Even though they demonstrate accurate surface reconstruction in Gaussian noise cases, the computational cost is high, and the denoising outcomes deteriorate with other noise types. Furthermore, a cascaded denoising framework is presented by the authors in \cite{wei2017tensor}. Their approach includes multi-scale tensor voting, vertex clustering step for detecting sharp features and a piece-wise fitting step for preserving the identified features.


\subsubsection*{Learning based}
Several studies employ deep networks for mesh denoising \cite{wang2016mesh,wang2019data,zhao2019normalnet,sarkar20183d}. 
The authors in \cite{wang2016mesh} suggest a data-driven method for mesh denoising that uses training sets of noisy objects. 
The objects are scanned by the same devices, thus facilitating the denoising of geometries with similar noise. 
The geometric features are reconstructed sufficiently.
Yet, their main limitation is that the reconstruction accuracy of important details relies heavily on whether they were initially included in the training set. 
In the same fashion, the authors in \cite{wang2019data} present a two-step ELM based denoising approach, where the first step performs coarse denoising and the second step recovers features. Several other research groups use CNNs working on voxelized versions of the geometry \cite{zhao2019normalnet}, or on images derived from local patches \cite{sarkar20183d}. However, our approach aims to be applied directly on the mesh nodes avoiding preprocessing, thus contributing to the field of geometric deep learning where the sampling of the latent space is nonuniform.  

\begin{table}
\label{table:notation_table}
  \centering
  \caption{Summary of Notations}
\begin{tabular}{r|l}
    \toprule

$\mathbf{v}_i$ & Vertex i\\
$\mathbf{c}_i$ & Centroid of face i\\
$f_i$ & Face i,$f_i=\{\mathbf{v}_{i1},\mathbf{v}_{i2},\mathbf{v}_{i3}\}$\\
$A_i$ & Face area\\
$n$ & Number of vertices\\
$n_f$ & Number of faces\\
$\mathcal{N}_i$ & Set of neighbouring vertices of vertex i\\
$\mathcal{N}_{f_i}$ & Set of neighbouring vertices of faces\\
$\mathbf{n}_{c_i}$ & Normal vector for centroid of face i, $\mathbf{n}_{c_i}\equiv\mathbf{n}_{f_i}$\\
$\mathbf{n}_{f_i}$ & Normal vector for face i, $\mathbf{n}_{c_i}\equiv\mathbf{n}_{f_i}$\\
$\delta_{n_1}$ & Rotation angle\\
$\mathbf{a}_{n_1}$ & Rotation axis\\
$\mathbf{a}_c$ & Arbitrary vector facilitating patch rotation\\
 $\mathbf{x}$  & Hidden layer output\\ 
$\mathbf{y}$  & Hidden layer output\\ 
$\mathbf{b}$  & Hidden layer bias\\ 
$\mathbf{z}$  & Autoencoder output vector\\ 
$\mathbf{W}_{E_{H_{1,2,3}}}$ & Weighting tensors for the encoder part  \\
$\mathbf{W}_{D_{H_{1,2}}}$ & Weighting tensors for the decoder part \\ $\mathbf{B}_{E_{H_{1,2,3}}}$ & Bias tensors for the encoder part  \\
$\mathbf{B}_{D_{H_{1,2}}}$ & Bias tensors for the decoder part\\
$L(\cdot)$  & Loss function\\ 
$s(\cdot)$  & Sigmoid function\\ 
$q_D(\mathbf{X})$, $q_\Phi(\mathbf{X})$  & Empirical distributions associated to $n$ training inputs\\ 
$D_{KL}$  & Kullback-Leibler Divergence \\ 
$\mathbf{H}(\mathbf{a},\mathbf{b})$  & Cross entropy loss\\ 
$\left[\mathbf{A}\dashline\mathbf{B}\right]$  &Concatenation of matrices $\mathbf{A}$ and $\mathbf{B}$\\ 
$\mathcal{L}$  &Evidence lower bound (ELBO) error\\
$\langle \mathbf{a} | \mathbf{b}  \rangle$ &Inner product of $\mathbf{a}$ and $\mathbf{b}$\\
$\bar{L}_p$ &Average edge length for 3D mesh \\
$N_B$ &Number of bilateral filter iterations \\
$N_V$ &Number of vertex update iterations\\
$\mathbf{1}$ & All ones vector [1\;1\;1]\\
    \bottomrule
  \end{tabular}
\end{table}

\section{Preliminaries}

\subsection{Preliminaries on deep autoencoders}
\label{subsection:preliminaries_autoencoders}
Deep autoencoders encompass a multi-layer neural network architecture where the hidden layers encode the input to a latent space 
and decode the latter to a reconstructed output. 
A deep autoencoder is composed of two symmetrical deep-belief networks \cite{hinton2009deep} that typically have three to five shallow layers for the encoding and the decoding part. 
The layers are Restricted Boltzmann Machines (RBMs). 
Variational autoencoders (VAE) \cite{kingma2013auto} assume that the input vectors are generated by some random process of an unobserved continuous random variable $\mathbf{z}$. 
The parameters of the VAE are estimated efficiently by the stochastic gradient variational Bayes framework \cite{kingma2013auto}. 
Furthermore, conditioning input vector $\mathbf{x}$ under label $c$ constitutes the basis of conditional variational autoencoder (CVAE) \cite{sohn2015learning}.

\begin{figure*}[t]
    \centering
    \includegraphics[width=\linewidth]{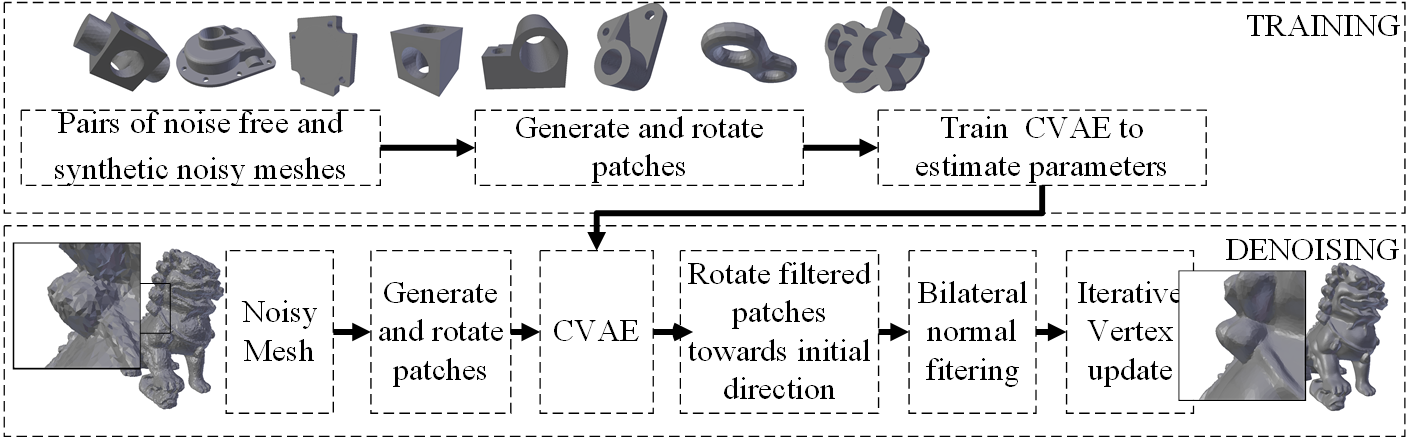}
    \caption{\small Pipeline of the proposed approach.
    }
    \label{fig:pipeline}
\end{figure*}
\begin{figure*}[t]
    \centering
    \includegraphics[width=\linewidth]{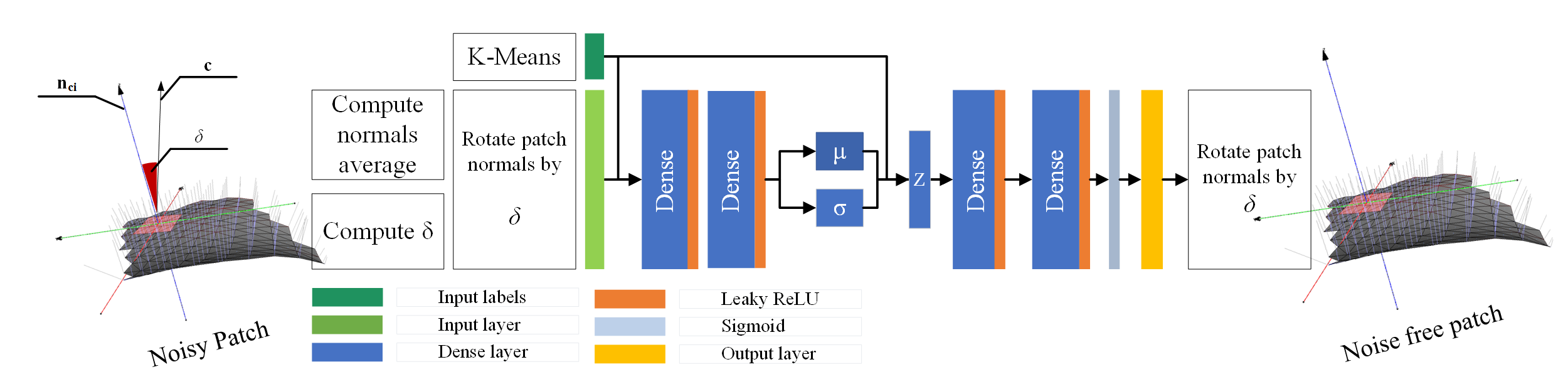}
    \caption{\small CVAE architecture and training scheme. Normal coordinates of noisy and noise-free patches are employed. Initially, they are properly rotated and labeled with k-means clustering.}
    \label{fig:architecture}
\end{figure*}

\subsection{Preliminaries on 3D meshes}
\label{subsection:meshes}

In this work, we focus on triangular meshes $\mathcal{M}$ with $n$ vertices $\mathbf{v}$ and $n_f$ faces $f$. Each vertex $\mathbf{v}_i$ is denoted by $\mathbf{v}_i = \left[x_i,\ y_i,\ z_i\right]^T, \ \forall \ i = 1,\cdots,n$.
Each $f_j$ face is a triangle that can be described by its centroid:
\begin{equation}
    \mathbf{c}_j =\nicefrac{\left(\mathbf{v}_{j1}+\mathbf{v}_{j2}+\mathbf{v}_{j3}\right)}{3}
\end{equation}
and its outward unit normal:
\begin{equation}
\mathbf{n}_{c_i} = \frac{\left(\mathbf{v}_{j2}-\mathbf{v}_{j1}\right) \times \left(\mathbf{v}_{j3}-\mathbf{v}_{i1}\right)}{\left\|\left(\mathbf{v}_{j2}-\mathbf{v}_{j1}\right) \times \left(\mathbf{v}_{j3}-\mathbf{v}_{j1}\right)\right\|}
\label{face_normals}
\end{equation}
where $\mathbf{v}_{j1}$, $\mathbf{v}_{j2}$ and $\mathbf{v}_{j3}$ are the position of the vertices that define face $f_j = \{\mathbf{v}_{j1} \ \mathbf{v}_{j2} \ \mathbf{v}_{j3} \}, \ \ \ \forall \ j = 1,\cdots,n_f$. 
 The first-ring area of a vertex $\mathbf{v}_i$ is defined as the neighborhood $\mathcal{N}_i$ in which the vertex $\mathbf{v}_i$ is connected to other vertices by only one edge (i.e., with topological degree equal to 1).

\section{Autoencoder architectures for 3D mesh denoising}
This section presents the mesh denoising pipeline. 
Training data were generated from meshes, distorted by noise, using the normal vectors corresponding to the 3D mesh faces. 
After training the autoencoder, the generated output vector is used for a normal-based vertex update\cite{zheng2011bilateral} of the mesh vertices. 

\subsection{Autoencoder architectures for mesh denoising}
\label{subsection:architectures}
This section presents the deep network architecture for mesh denoising. 
Specifically, a conditional variational autoencoder \cite{sohn2015learning} was employed, as illustrated in Figure \ref{fig:architecture}. 
A conditional Gaussian encoder with two dense layers is succeeded by a conditional Bernoulli decoder with two dense layers. 
Each dense layer is succeeded by a layer of leaky rectified linear units (ReLUs) and a dropout layer. 
We denote $\mathbf{X}$ as the input tensor, $\mathbf{Y}$ the corresponding labels, $\mathbf{X}_{in}=[\mathbf{Y}\dashline\mathbf{X}] $ the concatenation of $\mathbf{X}$ and $\mathbf{Y}$ and $\mathbf{X}_{out}$ the output of the autoencoder. The conditional Gaussian encoder is described as follows:

\begin{align}
\mathbf{X}_{in} &= [\mathbf{Y}\dashline\mathbf{X}] \\
\mathbf{Y}_{E_{H_1}}  &= \max\left(0,\mathbf{X}_{in} \mathbf{W}_{E_{H_1}}+\mathbf{B}_{E_{H_1}}\right)\\
\mathbf{Y}_{E_{H_2}}  &= \max\left(0,\mathbf{Y}_{E_{H_1}}  \mathbf{W}_{E_{H_2}}+\mathbf{B}_{E_{H_2}}\right)\\
\mathbf{Y}  &= \max\left(0,\mathbf{Y}_{E_{H_2}}   \mathbf{W}_{E_{H_3}}+\mathbf{B}_{E_{H_3}}\right)\\
\mathbf{\mu} &= [\mathbf{y}_1\;\mathbf{y}_2 ] ,\;  \mathbf{\sigma} = [\mathbf{y}_3\;\mathbf{y}_4 ]\\
\mathbf{Z}  &= \mu  + \sigma\cdot\mathcal{N}(0,1)
\end{align}
Subsequently, the conditional Bernoulli decoder is described as follows:
\begin{align}
\mathbf{Z} &= \left[\mathbf{Y}\dashline\mathbf{Z}\right]\\
\mathbf{Y}_{D_{H_1}}  &= \max\left(0,\mathbf{Z}  \mathbf{W}_{D_{H_1}}+\mathbf{B}_{D_{H_1}}\right)\\
\mathbf{Y}_{D_{H_2}}  &= \max\left(0,\mathbf{Y}_{D_{H_1}}  \mathbf{W}_{D_{H_2}}+\mathbf{B}_{D_{H_2}}\right)\\
\mathbf{Y}_{D_{H_2}}  &= \max\left(0,\mathbf{Y}_{D_{H_1}}  \mathbf{W}_{D_{H_2}}+\mathbf{B}_{D_{H_2}}\right)\\
\mathbf{X}_{out}  &= \sigma\left(\mathbf{Y}_{D_{H_1}} \mathbf{W}_{D_{H_2}}+\mathbf{B}_{D_{H_2}}\right)
\end{align}
where $\mathbf{W}_{E_{H_{1,2,3}}}$,$\mathbf{W}_{D_{H_{1,2}}}$ and $\mathbf{B}_{E_{H_{1,2,3}}}$,$\mathbf{B}_{D_{H_{1,2}}}$ are weighting and bias tensors respectively. 
For the training of the autoencoder architecture the evidence lower bound (ELBO) error is computed:
\begin{equation}
\mathcal{L}=H(\mathbf{X}_{in},\mathbf{X}_{out})+D_{KL}(\mathbf{\mu}||\mathbf{\sigma})
\end{equation}
where $H(\mathbf{X}_{in},\mathbf{X}_{out})$ is the cross entropy and  $D_{KL}(\mathbf{\mu}||\mathbf{\sigma})$ the Kullback-Leibler divergence.

\subsection{Training, denoising and post-processing}
This section describes the autoencoder training and denoising pipeline, depicted in Figures \ref{fig:pipeline} and \ref{fig:architecture}, and the patch descriptor utilized in the proposed scheme.

\subsubsection{Patch descriptor}
\begin{figure}
    \begin{center}
    \includegraphics[width=\linewidth]{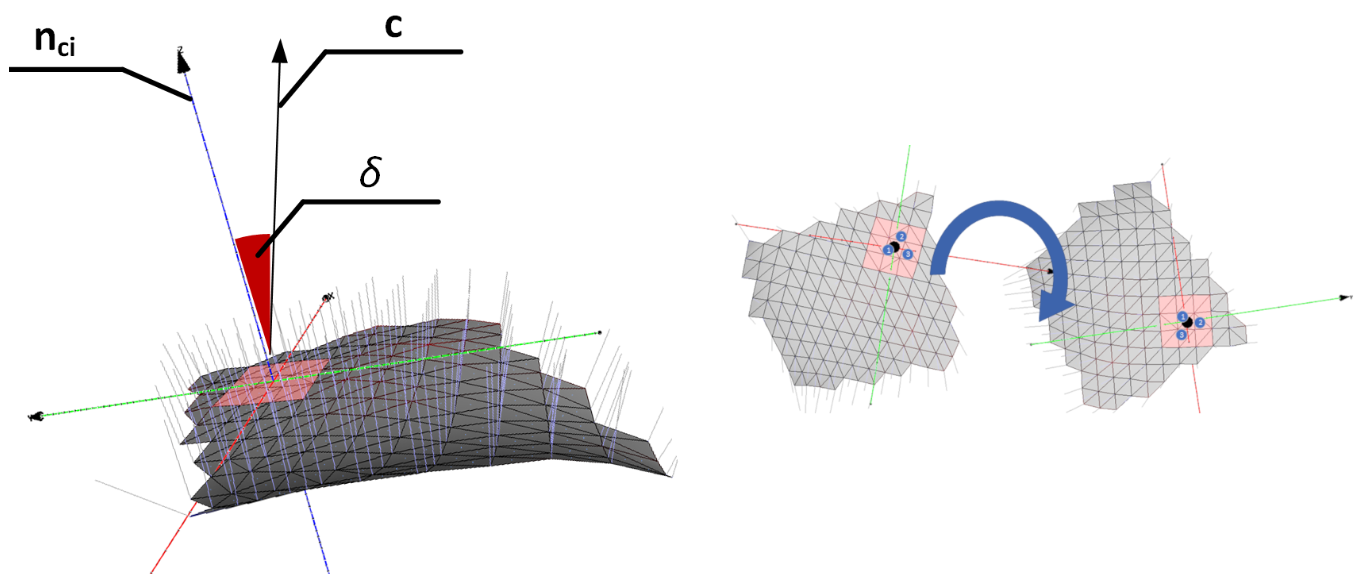}
     \begin{subfigure}{0.49\linewidth}
    \caption{}
    \end{subfigure}%
    \begin{subfigure}{0.49\linewidth}
    \caption{}
    \end{subfigure}
  \end{center}
\caption{\small Example of a patch neighbourhood (n = 17), highlighted in orange color, of the face highlighted in cyan color. The normals of the faces are also apparent. }
\label{fig:patch}
\end{figure}

The patch descriptor aims to constrain the latent space, to allow for efficient training of the autoencoder. 
Each patch $\mathcal{P}_i$ is comprised of $N$ topological neighbours of face $f_i$. 
Neighbouring faces are sorted using the distance of the face centroid $c_i$ to the face centroid $c_j$, where $f_j$ is a neighbouring face belonging to patch $\mathcal{P}_i$, $f_j \in \mathcal{P}_i$. 
A schematic visualization of the patch descriptor is depicted in Figure \ref{fig:patch}. 
Assuming a local coordinate system (Figure \ref{fig:patch}a), arranging faces constrains the latent space in the $\mathbf{z}$-axis. 
To further constrain the latent space across $\mathbf{x}$ and $\mathbf{y}$ axes the patch is rotated by angle $\delta_{n_i}$ around rotation axis $\mathbf{a}_{n_1}$ so that: 
\begin{equation}
    \mathbf{n}_f=\frac{1}{N}\sum_{i \in \mathcal{N}_i } A_i\cdot\mathbf{n}_{c_i}=\mathbf{a}_c
    \label{eq:rotation}
\end{equation}
where $\mathbf{a}_c$ is a known arbitrarily defined vector.
The motivation behind rotating each patch towards the same direction is that it allows efficient training with smaller training sets. 
Otherwise, we would have to include patches with every possible direction of normals to the training dataset, resulting in large datasets.

\subsubsection{Training} 

\begin{figure}
    \begin{center}
    \includegraphics[width=\linewidth]{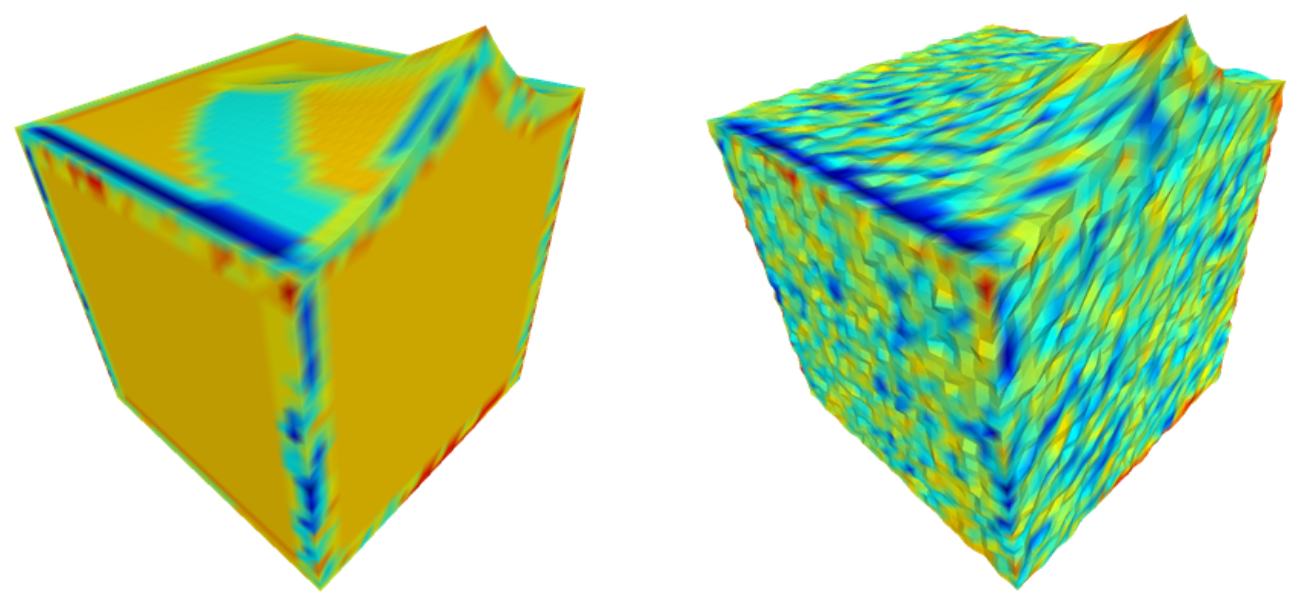}
    \begin{subfigure}{0.49\linewidth}
    \caption{}
    \end{subfigure}%
    \begin{subfigure}{0.49\linewidth}
    \caption{}
    \end{subfigure}
    
      \end{center}
    \caption{\small Visualization of the K-means clustering for a (a) noise-free, and (b) a noisy mesh of the same model.}
    \label{fig:kmeans_clustering}
\end{figure}

The training of the deep network is schematically presented in Figures \ref{fig:pipeline} and \ref{fig:architecture}.
The training set contains pairs of noisy and noise-free patches comprised of $N$ neighbouring faces. The corresponding face normals are rotated by $\delta_{n_i}$ around rotation axis $\mathbf{a}_{n_1}$. For the definition of the rotation axis the normals of the noisy patch are used as reference.
In order to generate labels for the training set, we perform K-means clustering defining the group centroids for $K$ clusters. 
The motivation behind applying K-means clustering is that it divides the dataset into groups of patches with high curvature, low curvature, flat areas, and features i.e corners. Thus, different models are trained for each category. Figure \ref{fig:kmeans_clustering} presents an example of a 3D mesh and its corresponding noisy version. The K-means clustering of the different surface categories is depicted using different color per different cluster.
The coordinates of the normalized normal vectors $\mathbf{n}_{c_i}$, that comprise patch $\mathcal{P}_i$, range in $[-1,1]$. 
They are transformed to range in $[0,1]$ by the following equation: 
\begin{equation}
    \mathbf{n}_{c_i}'=2\cdot\mathbf{n}_{c_i}-\mathbf{1}
    \label{eq:normal_normalized}
\end{equation}
Subsequently, the matrix $\mathbf{N}_i^{3 \times (n+1)}$, consisting of vectors $\mathbf{n}_{c_i}'$ is reshaped to $\mathbf{Z}_i^{3(n+1) \times 1}$. Finally, training is performed with Adam optimizer.

\subsubsection{Denoising} 
\label{subsection:denoising}
The denoising process is visualized in Figure \ref{fig:pipeline}. 
To use the trained autoencoder for denoising, patches are formed on the noisy mesh. 
For each patch the average normal $\mathbf{n}_f$ is extracted and the patch is rotated by $\delta_k$ so that $\mathbf{n}_f$ is co-directional to $\mathbf{a}_c$, to form the input matrix $\mathbf{Z}_i^{3(n+1) \times 1}$. 
After the autoencoder has generated the filtered output $\mathbf{Z'}_i^{3(n+1) \times 1}$, they are reshaped back to the original form $\mathbf{Z'}_i^{3 \times (n+1)}$. 
The exported filtered normal vector for patch $\mathcal{P}_i$ is the first column of $\mathbf{Z'}$, and more specifically:
\begin{equation}
\mathbf{\hat{n}}_{c_i}^{3\times1} =\mathbf{Z'}_i[:,0]
\end{equation}
Finally, each patch is rotated by the opposite angle $-\delta_{n_i}$ and the same axis $\mathbf{a}_{n_i}$, that they were rotated with in the first place. 

\subsubsection{Post-processing}
\label{subsection:postprocessing}
As a final post processing step, we use the bilateral filtering approach according to \cite{zheng2011bilateral}:
\begin{equation}
\mathbf{\hat{n}}_c =  \frac{\sum_{f_j \in \mathcal{P}_i} A_j \mathbf{W}_{{1}_{ij}} \mathbf{W}_{{2}_{ij}} \mathbf{n}_{cj}} {\|A_j \mathbf{W}_{{1}_{ij}} \mathbf{W}_{{2}_{ij}} \mathbf{n}_{cj}\|_2}
\label{bilateral_filtering}
\end{equation}
\begin{equation}
\small
\mathbf{W}_{{1}_{ij}} =
\text{exp}(\frac{-\begin{Vmatrix}\mathbf{c}_{i}-\mathbf{c}_{j}\end{Vmatrix}^{2}}{2\sigma^{2}_{1}} ), \ \
\mathbf{W}_{{2}_{ij}} =  
\text{exp}(\frac{-\begin{Vmatrix}\mathbf{n}_{ci}-\mathbf{n}_{cj}\end{Vmatrix}^{2}}{2\sigma^{2}_{2}} )
\label{Ws}
\end{equation}
where $A_j$ represents the area of face $f_j$.
Finally, the denoised normals $\mathbf{\hat{n}}_c$ are used to update the vertices according to \cite{sun2007fast}:
\begin{equation}
\mathbf{v}_{i} = \mathbf{v}_{i} + \frac{ \sum_{\mathbf{c}_j \in \mathcal{N}_i} \mathbf{\hat{n}}_{cj} ( \langle \mathbf{\hat{n}}_{cj} | (\mathbf{c}^{}_{j} - \mathbf{v}^{}_{i}) \rangle )}{|\mathcal{N}_i|} 
\label{vt}
\end{equation}
where $\mathcal{N}_i$ represents the first-ring area of a vertex $\mathbf{v}_i$.
At this point, it is significant to clarify that we always use the same values for each model, without searching for the ideal parameters per model. More specifically, $\sigma_2=0.15$ and for the estimation of $\sigma_1$, we use the following equation: 
\begin{equation}
    \sigma_{1_{ij}} = \frac{\sum_{\forall \mathbf{c}_j \in \mathcal{N}_i} \|\mathbf{c}_i - \mathbf{c}_j\|_2^2}{|\mathcal{N}_i|}
\end{equation} 
as proposed by \cite{zheng2011bilateral}.

\color{annotated}
For the rest of the paper, we define $N_B$ the number of bilateral filtering iterations and $N_V$ the number of vertex update iterations that are performed.  
Furthermore, Gaussian noise will be defined as $N \sim (\mu,\;\sigma)$,
where $\mu$ is the mean value, $\sigma=\beta\cdot \bar{L}_p$ is the standard deviation, $\bar{L}_p$ is the average edge length and $\beta$ a scalar value.

\color{black}

$N_V$ is set to $N_V=20$, while $N_B$ depends on the noise level. Experimental evaluation showed that for $N \sim (0,\;0.1\cdot \bar{L}_p)$, $N_B=1$ allows fine-tuning by removing small artifacts. More iterations increase the computational cost, without any additional benefit. Further elaboration, is presented in subsection \ref{subsection:experimental_setup}. 
Algorithm \ref{al:pipeline_AOI} summarizes the steps of the proposed method. 
Even though this training process is time-consuming, it takes place only once. Also, in comparison with other data-driven methods \cite{wang2016mesh}, the training process is faster, due to the smaller required dataset size.

\begin{algorithm}
    \small
    \SetKwInOut{Input}{Input}
    \SetKwInOut{Output}{Output}
    \tcp{Training Process}
    \Input{Noisy Dataset of meshes $\mathcal{M}_p$, Original Dataset of meshes $\mathcal{M'}_p \ \in \ \mathbb{R}^{n_{f_p} \times 3k}, \ \ \forall \ p  \ \in$ dataset;}
    \Output{Weights $\mathbf{W}_{E_{H_i}},\mathbf{W}_{D_{H_i}},i\in[1,2,3]$;}
    Estimate centroid normals via Eq. \eqref{face_normals}; \\
    \For{$i = 1, \cdot, n_{f_p}$}{
    Rotate the normals by angle $\delta_{n_i}$ around rotation axis $\mathbf{a}_{n_i}$ via Eq. \eqref{eq:rotation}; \\
    Normalize the normal vectors in a range of [0,1] via Eq. \eqref{eq:normal_normalized}; \\
    }
    Compute labels $\mathbf{Y}$\\
    Reshape and create training set $\mathbf{Z}_i^{3(n+1) \times 1}$ and $\mathbf{Z'}_i^{3(n+1) \times 1}$;\\
    $\mathbf{W}_{E_{H_i}},\mathbf{W}_{D_{H_i}},i\in[1,2,3] = \text{train}(\mathbf{Z}_i,\mathbf{Z'}_i);$
    
   \tcp{Denoising Process}
       \Input{Noisy model $\mathcal{M} \ \in \ \mathbb{R}^{{n}_f \times 3k}$;}
       \Output{Denoised model $\mathcal{\tilde{M}} \ \in \ \mathbb{R}^{{n}_f \times 3k}$;}
       Estimate centroid normals via Eq. \eqref{face_normals}; \\
       \For{$i = 1, \cdots, \tilde{n}_f$}{
       Rotate the normals by angle $\delta_{n_i}$ around rotation axis $\mathbf{a}_{n_i}$ via Eq. \eqref{eq:rotation}; \\
    Normalize the normal vectors in a range of [0,1] via Eq. \eqref{eq:normal_normalized}; \\
    }
    Reshape and create the input $\mathbf{Z}_i^{3(n+1) \times 1}$;\\
   $\mathbf{{\bar{n}}}_{\mathbf{c}i} = \text{denoising}(\mathbf{Z}_i,\mathbf{W}_{E_{H_i}},\mathbf{W}_{D_{H_i}},i\in[1,2,3])$; \\
       
       Post processing for fine-tuning using bilateral filtering via Eqs. \eqref{bilateral_filtering}-\eqref{Ws}; \\
       Reconstruction of the final denoised model using vertices updating via Eq. \eqref{vt};
    \caption{Data Training and Denoising of 3D Meshes} \label{al:pipeline_AOI}
    
\end{algorithm}


\section{Experimental analysis and simulation results}

\subsection{Experimental setup and training}
\label{subsection:experimental_setup}

\color{annotated}
Two different datasets are examined. The first includes meshes originating from the shape repository of the AIM@Shape project \cite{falcidieno2004aim} with synthetic Gaussian noise.
The second utilizes Kinect 2 scans of 3D printed objects provided by Wang et al. \cite{wang2016mesh}. 
The latter provides noisy scanned outcomes along with ground truth models.
\color{black}
To test the denoising capability of our method\footnote{https://github.com/snousias/fast-mesh-denoising}, we compared our results to guided mesh normal filtering \cite{zhang2015guided}, bilateral normal filtering \cite{zheng2011bilateral}, $L_0$ minimization mesh denoising \cite{he2013mesh}, fast and effective mesh denoising \cite{sun2007fast}, mesh denoising via cascaded normal regression \cite{wang2016mesh} and feature preserving mesh denoising based on graph spectral processing\cite{arvanitis2019feature}. 

As an additional comparison, the CVAE part of our pipeline was replaced with traditional autoencoders, referred to as AE. For the latter, a 5-layer deep autoencoder was employed with $N=[256,128,64,128,256]$ the number of neurons for each layer. An element-wise sigmoid operation succeeds each layer, trained with a mean square error loss function.

\subsubsection{Synthetic Gaussian Noise }
Eight meshes were selected for the training of the autoencoder architecture, comprising in total of 1,977,740 patches. Noisy meshes were synthesized by adding Gaussian noise $N \sim (0,\;0.1\cdot \bar{L}_p)$ co-directional to each vertex normal. 1,977,740 training pairs of noisy and noise-free rotated patches were utilized for the training of the autoencoder. 
Two configurations were tested for patch size, $n=8$ and $n=20$ neighbours. $K=200$ was selected for the K-means clustering of the CVAE. 
Training was performed with an Adam optimizer with $\beta_1=0.9$, $\beta_2=0.999$ and $\epsilon=1e-8$. The training took place for 100 epochs, utilizing an NVIDIA GeForce GTX 1080 graphics card with 8GB VRAM and compute capability $6.1$. 
For the bilateral filtering, we execute only $N_B=1$ iteration, while for the vertex update operation, we execute $N_B=20$ iterations.
Experimental evaluation showed that for noise level up to $N \sim (0,\;0.1\cdot \bar{L}_p)$ a single bilateral iteration adequately performs fine-tuning, by removing small artifacts, while more iterations increase the computational cost without any additional benefit. 

\subsubsection{Kinect scans}

Kinect scans were selected from the dataset provided by Wang et al. \cite{wang2016mesh} in order to form 927541 training examples in total. The trained CVAE model was employed to denoise a noisy Kinect scanned model excluded from the training set. The observed noise level of the Kinect scans was computed to $N \sim (-0.18,\;1.4\cdot \bar{L}_p)$, while Figure \ref{fig:kinect} presents the denoising outcome. 
Furthermore, different settings were tested to evaluate optimal patch size and number of bilateral iterations. 
Patch size ranged in $n=8,20,40,60$, the number of clusters in $K=6,10,50,100,200$ and the number of bilateral filtering iterations $N_B$ in $N_B=0,1,4,8$.

\subsubsection{Hyper-parameter optimization}
\begin{figure}

    \centering
\includegraphics[width=\linewidth]{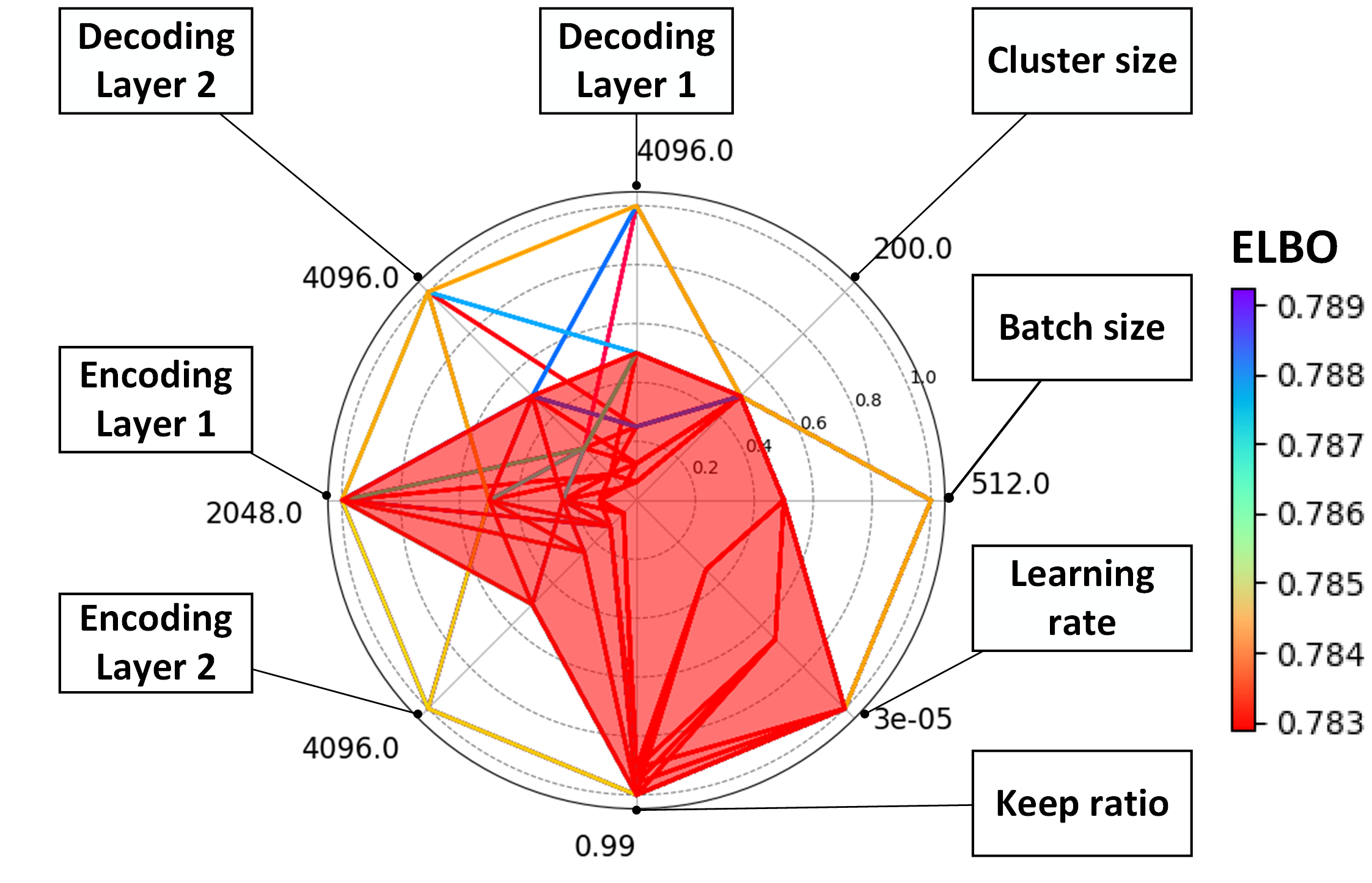}
    \caption{Hyper-parameter optimization radar chart}
    \label{fig:HPO_f}
\end{figure}

To define the number of nodes for each layer we performed hyper-parameter optimization. For encoding layers $E_1$, $E_2$ and decoding layers $D_1$, $D_2$ the number of nodes ranged in $N_{E_1},N_{E_2},N_{D_1},N_{D_2} \in [ 256, 512, 1024, 2048, 4096]$. The number of clusters was set equal to $K=200$, the patch size $n=20$, the learning rate ranged in $lr\in[1e-05,2e-05,3e-05]$, and the keep ratio ranged in $kr \in [0.90 , 0.95, 0.99]$.
We computed the set of parameters that exhibit higher performance, in terms of lower ELBO loss, as shown in Fig. \ref{fig:HPO_f}: i) Batch size equals $256$ ii) keep ratio $kr=0.99$, iii) learning rate $lr=3e-05$,iv) decrease ratio $dr=0.998$, v) $N_{E_1}=2048$, vi) $N_{E_2}=2048$, vii) $N_{D_1}=2048$ and viii) $N_{D_2}=2048$.

\subsubsection{Evaluation models and metrics}
The quality of the reconstructed results is evaluated using a) the Hausdorff distance (HD) which represents the average one-side distance between the reconstructed and the original 3D mesh, b) the metric $\alpha$ which represents the average angle difference between the normals of the ground truth and the reconstructed model c) visualizations which present in different colors the absolute difference between the reconstructed and original meshes.

\begin{table*}
\caption { Hausdorff distance and metric $\alpha$ (in degree) using disparate approaches of initialization and deep architectures. }
\begin{center}
\resizebox{\linewidth}{!}{
\begin{tabular}{|p{3cm}||c|c|c|c|c|c|c|}
\hline
                     & \textbf{Bilateral \cite{zheng2011bilateral}} & \textbf{Guided\cite{zhang2015guided}} & \textbf{Fast n Effective\cite{sun2007fast}} & \textbf{L0 min \cite{he2013mesh}} &
                     \textbf{Feature aware \cite{arvanitis2019feature}} &
                     \textbf{Cascaded \cite{wang2016mesh}} & \textbf{CVAE\_20\_pp} \\ \hline
\textbf{Carter (100000 F)}  & 7.934 / 0.507 & 9.16 / 0.602 & 11.519 / 0.89 & 7.253 / 0.363 & 7.209 / 0.56 & 8.43 / 0.621 &5.955 / 0.457 \\ \hline
\textbf{Pulley (100000 F)}      & \textbf{5.476} / 0.365 & 7.984 / 0.455 & 8.573 / 0.627 & 6.341 / 0.255 & \textbf{3.786} / 0.827 & 6.78 / 0.440 & \textbf{3.591} / 0.305\\ \hline
\textbf{Screwdriver (54000 F)} & 4.191 / 0.003 & 4.654 / 0.004 & 6.389 / 0.006 & 4.829 / 0.003 & 2.58 / 0.004 & 3.93 / 0.003 & 3.161 / 0.003 \\ \hline
\end{tabular}
}
\end{center}

\label{table:AE2CVAEHD}
\end{table*}

\begin{table*}
\vspace{-1.8em}
\begin{center}
\resizebox{\linewidth}{!}{
\begin{tabular}{|p{3cm}||c|c|c|c||c|c|c|c|}
\hline
\multicolumn{1}{|l||}{} & \multicolumn{4}{c||}{\textbf{AE}}                                                    & \multicolumn{4}{c|}{\textbf{CVAE}}                                                          \\ \hline
                       & \textbf{AE\_8\_no} & \textbf{AE\_8\_pp} & \textbf{AE\_20\_no} & \textbf{AE\_20\_pp} & \textbf{CVAE\_8\_no} & \textbf{CVAE\_8\_pp} & \textbf{CVAE\_20\_no} & \textbf{CVAE\_20\_pp} \\ \hline
\textbf{Carter (100000 F)}& 5.345 / 0.343      & 5.206 / 0.406      & 5.159 / 0.471       & 5.847 / 0.523       & 5.531 / 0.341        & 5.364 / 0.39         & 5.282 / 0.405         & 5.955 / 0.457         \\ \hline
\textbf{Pulley (100000 F)}& 4.359 / 0.178      & 3.371 / 0.219      & 2.917 / 0.205       & 2.995 / 0.257       & 4.562 / 0.248        & 3.672 / 0.254        & 3.372 / 0.278         & 3.591 / 0.305         \\ \hline
\textbf{Screwdriver (54000 F)}& 4.179 / 0.003      & 2.923 / 0.003      & 3.315 / 0.004       & 3.113 / 0.004       & 4.112 / 0.002        & 2.946 / 0.003        & 3.347 / 0.003         & 3.161 / 0.003         \\ \hline
\end{tabular}

}
\end{center}
\vspace{-1.0em}
\end{table*}

\begin{table*}
\caption{Execution time,measured in seconds, for presented approaches and noise level $N \sim (0,\;0.1\cdot \bar{L}_p)$}
\centering
\resizebox{\linewidth}{!}{
\begin{tabular}{|l|l|l|l|l|l|l|l|}
\hline
\textbf{} & 
\textbf{Bilateral} & 
\textbf{Guided} & 
\textbf{Fast} & 
\textbf{L0} & 
\textbf{Feature} & 
\textbf{CVAE 20 pp} & 
\textbf{CVAE 8 pp} \\ 
\textbf{}& 
\textbf{normal\cite{zheng2011bilateral}} & 
\textbf{normal\cite{zhang2015guided}} & 
\textbf{effective\cite{sun2007fast}} & 
\textbf{min\cite{he2013mesh}} & 
\textbf{aware\cite{arvanitis2019feature}} & 
\textbf{} & 
\textbf{} \\ \hline
\textbf{Sculpt (3669V,7342F)} & 
0.1082 & 
0.6465 & 
0.0591 & 
3.5884 & 
0.26743 &
0.0772 & 
0.0754 \\ \hline
\textbf{Trimmed star (5192V, 10384F)} & 
0.1529 & 
0.9843 & 
0.0869 & 
4.2748 & 
0.41393 &
0.0995 & 
0.0995 \\ \hline
\textbf{Rocker Arm (9413V,18826F)} & 
\textbf{0.3242} & 
2.0561 & 
0.1804 & 
11.1609 & 
\textbf{1.10021} &
\textbf{0.1642} & 
\textbf{0.1617}\\ \hline
\textbf{Chinese Lion (50000 V, 100000F)} & 
2.0508 & 
21.6360 & 
1.5792 & 
110.6100 & 
16.24114 &
0.9872 & 
0.9624 \\ \hline
\textbf{Gear (250000V,500000F)} & 
8.5630 & 
221.1910 & 
5.7120 & 
2512.2500 & 
180.77456 &
3.8858 & 
3.7505 \\ \hline
\end{tabular}
}
\label{table:exec}
\end{table*}

\begin{figure*}
\centering
\includegraphics[width=\linewidth]{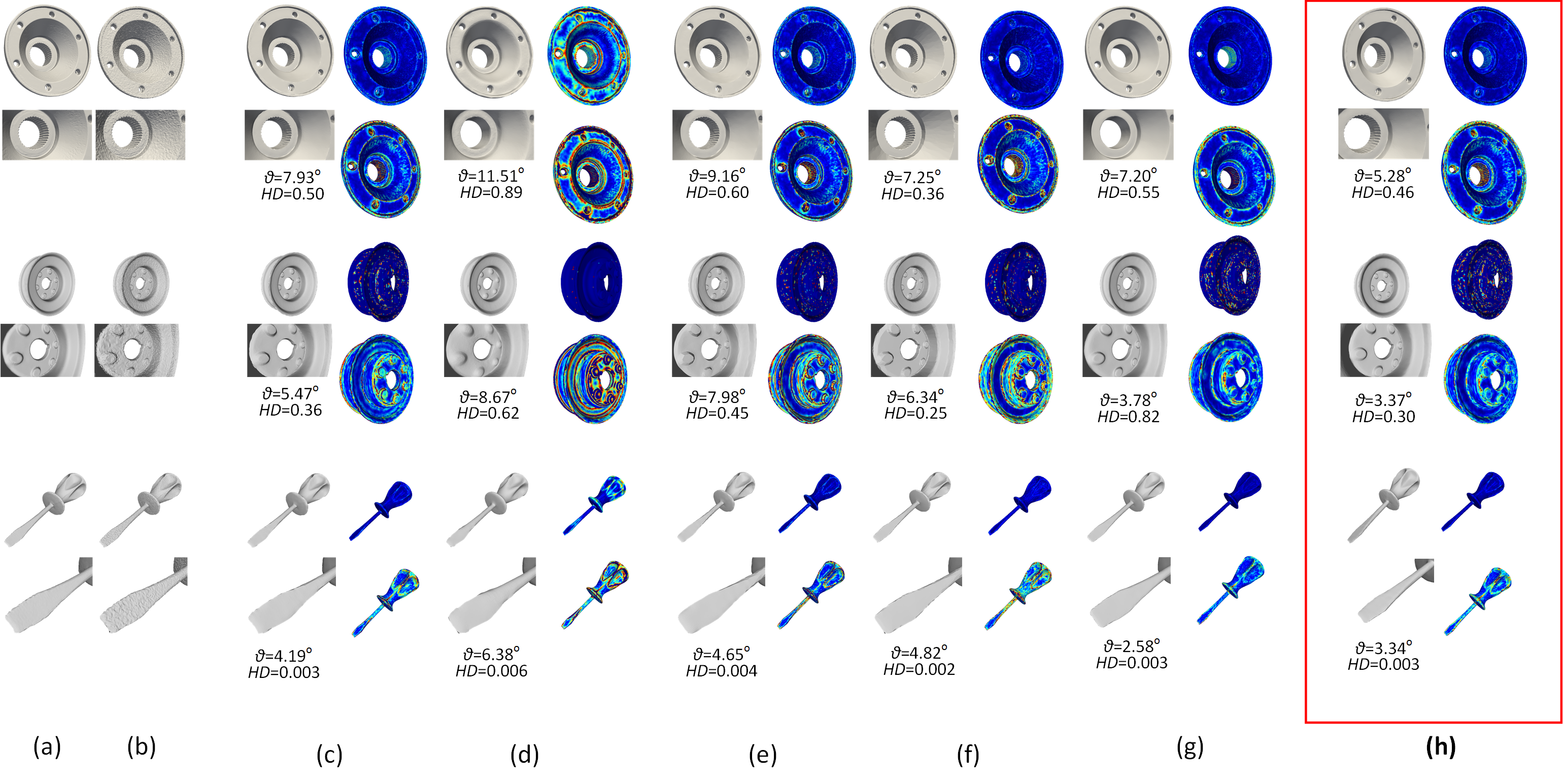}
\caption{\small Denoising results and normal angle difference visualization between reconstructed and original 3D model with Gaussian noise $N \sim (0,\;0.1\cdot \bar{L}_p)$. For each model absolute distance (upper colored mesh) and theta distribution (lower colored mesh) are presented. 
(a) original mesh, (b) noisy mesh, (c) fast and effective \cite{sun2007fast}, (d) bilateral normal filtering \cite{zheng2011bilateral}, (e) $L_0$ minimization \cite{he2013mesh}, (f) guided normal filtering \cite{zhang2015guided},(g) Cascaded mesh denoising \textbf{(h) our approach}.
}
\label{fig:denoising_d1}
\end{figure*}

\begin{figure*}
\centering
\includegraphics[width=\linewidth]{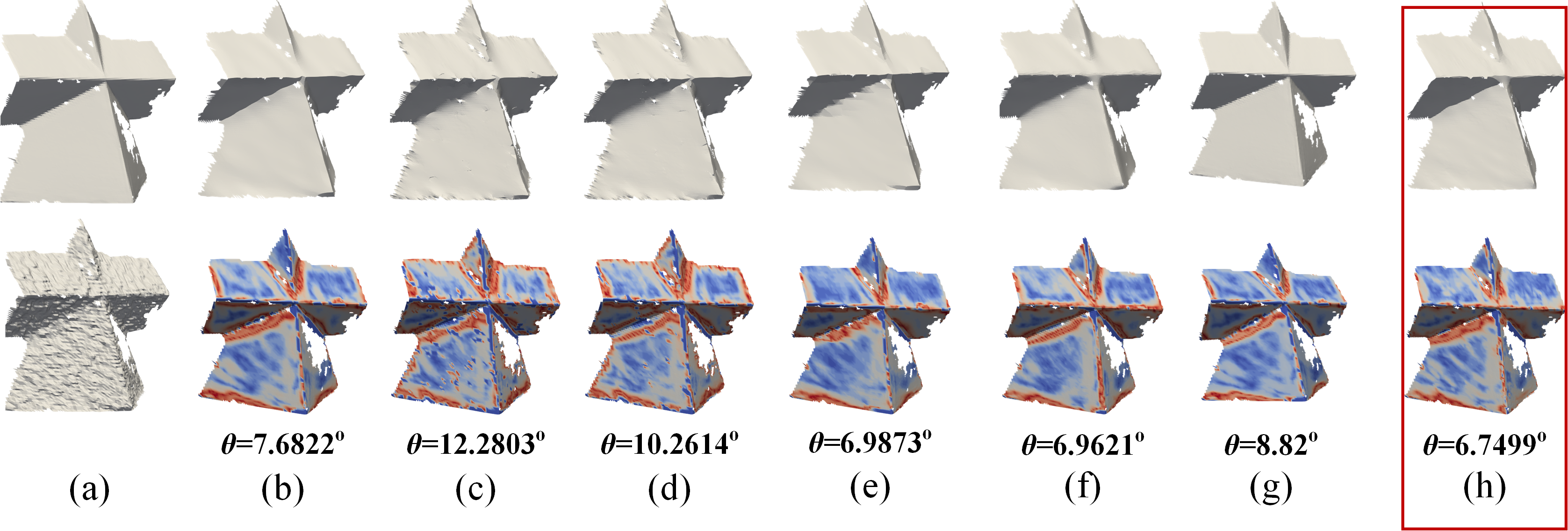}
\caption{
\small Denoising results and normal angle difference visualization between reconstructed and original Kinect scan with noise level $N \sim (-0.18,\;1.4\cdot \bar{L}_p)$.
(a) original mesh (up) and noisy mesh (down), (b) fast and effective \cite{sun2007fast}, (c) bilateral normal filtering \cite{zheng2011bilateral}, (d) $L_0$ minimization \cite{he2013mesh}, (e) guided normal filtering \cite{zhang2015guided}, (f) Cascaded mesh denoising, (g) Feature aware denoising\cite{arvanitis2019feature} and \textbf{(h) our approach.}}
\label{fig:kinect}
\end{figure*}

\begin{figure}[h]
    \centering
    \includegraphics[width=\linewidth]{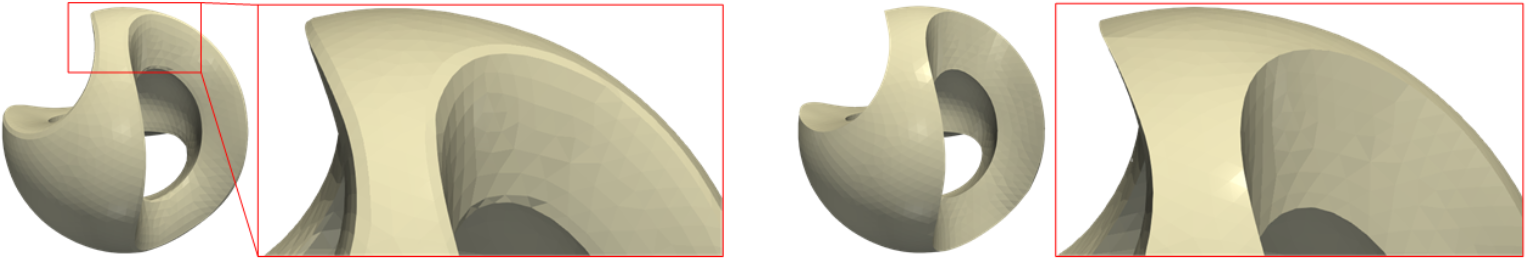}
    \begin{subfigure}{0.49\linewidth}
    \caption{}
    \end{subfigure}%
    \begin{subfigure}{0.49\linewidth}
    \caption{}
    \end{subfigure}
    \caption{\small Denoising results using: (a) AE, (b) CVAE. }
    \label{fig:AE_vs_CVAE}
\end{figure}

\subsection{Mesh denoising studies}
\subsubsection{Evaluation of reconstructed models}
Table \ref{table:AE2CVAEHD} presents the Hausdorff distance and the mean angular difference $\alpha$ of the face normals between original and denoised 3D models correspondingly. In these tables, we use a variety of different initialization approaches and architectures\cite{zheng2011bilateral,he2013mesh,zhang2015guided,arvanitis2019feature,wang2016mesh,sun2007fast}. More specifically, we deployed two different deep architectures (i.e., AE and CVAE), in two different patch sizes (i.e., with 8 and 20 nearest neighbours (nn)), and with, or without \textbf{post-processing step (pp)}. 
As we can observe, the best performance depends on the model, and none of these approaches is universally the best. Nevertheless, in most of the cases, the CVAE using 8 nearest neighbours seems to have the most stable behaviour. Comparing the reconstructed meshes provided by AE and CVAE, we notice that simple AE gives a smoothed result to the object's surface, but it negatively affects the preservation of features. On the other hand, CVAE achieves the accurate reconstruction of geometrical features, but the surface of flat areas contains artifacts, as shown in Figure \ref{fig:AE_vs_CVAE}. However, this is a problem efficiently tackled by the post-processing step.

Figure \ref{fig:denoising_d1} presents a visual comparison of the reconstructed models. In this figure, we also provide enlarged details as well as the $\alpha$ metric for easier evaluation. Additionally, Figure \ref{fig:denoising_d1} illustrates a visualization of the absolute distance and the theta metric between the original and the reconstructed model for each vertex of the meshes. The lowest value (dark blue) denotes that the compared vertices have the same position, in the 3D coordinate system, while a high value (dark red) of the absolute distance denotes that the vertices exhibit high error. Figure \ref{fig:kinect} presents the denoising result for the Kinect2 scanned models. Our approach accomplishes a lower theta mean value yielding equivalent results with other established data-driven approaches\cite{wang2016mesh}.

\subsubsection{Impact of patch size, number of clusters and filter parameters}
\begin{figure}
    \centering
    \includegraphics[width=\linewidth]{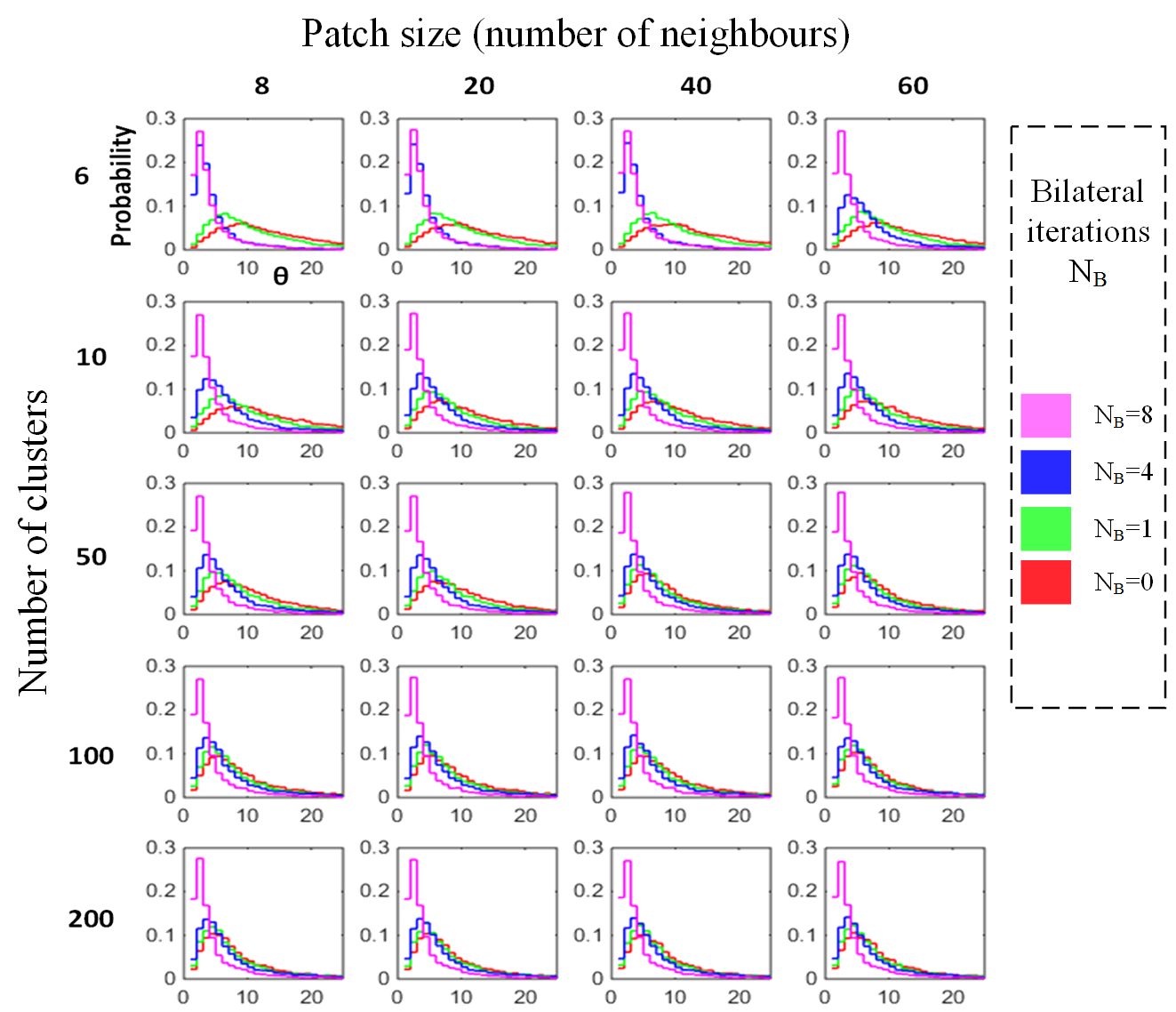}
    \caption{\small Evaluation of theta distribution for different settings of the number of clusters, the number of patch neighbours and bilateral filter iterations.}
    \label{fig:hyperparameters}
\end{figure}

Figure \ref{fig:hyperparameters} presents the theta distribution for different settings of selected number of clusters, patch size and bilateral iterations. Purple lines correspond to $N_B=8$ iterations, blue lines to $N_B=4$ iterations, green lines to $N_B=1$ iterations and red lines to $N_B=0$ iterations (no post processing). As we can observe, $8$ iterations significantly improve the result for $N \sim (-0.18,\;1.4\cdot \bar{L}_p)$ noise.

\subsection{Computational complexity evaluation}
This subsection presents a comparison of our approach with other methods in terms of computational complexity. To facilitate the performance evaluation we used an open-source implementation in C++ of state-of-the-art methods \cite{zheng2011bilateral,zhang2015guided,sun2007fast,he2013mesh} available in \cite{zhang2015meshgit}. To be more specific, the execution was totally performed in C++.
For our approach, the autoencoder part of our pipeline is executed in Python TensorFlow, the denoised normal rotation, bilateral normal filtering and vertex update parts in C++. All the evaluation studies took place in a Intel(R) Core(TM) i7-4790 CPU @ 3.60Hz with 32GB of RAM.

\begin{figure}
    \centering
    \includegraphics[width=\linewidth]{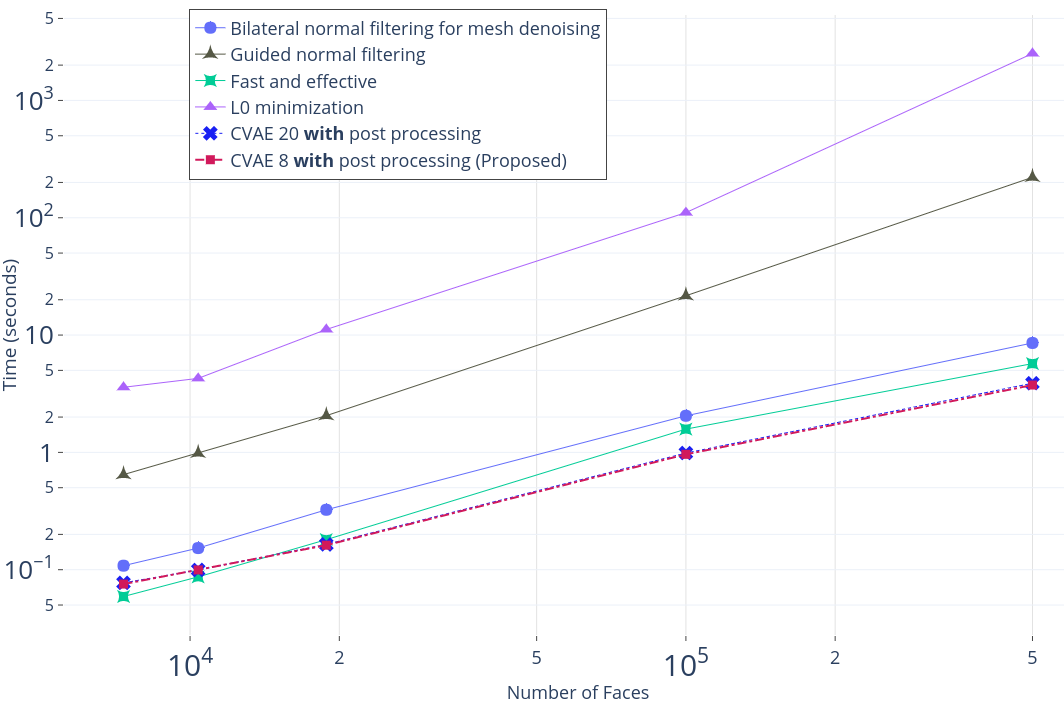}
    \caption{\small Performance evaluation.}
    \label{fig:performance}
\end{figure}

\color{annotated}
As Figure \ref{fig:performance} and Table \ref{table:exec} show, our method is much faster than $L_0$ minimization \cite{he2013mesh} and Guided Normal Filtering \cite{zhang2015guided} and traditional bilateral normal filtering \cite{zheng2011bilateral}.
Compared to fast and effective mesh denoising \cite{sun2007fast}, our method is slower in small models but becomes faster as the number of faces increases.
Execution time measurements presented in Table \ref{table:exec} were computed as the mean value of 10 repetitions. In the case of Rocker Arm counting 18826 faces, our method outperforms all the other approaches. We attribute this observation to the autoencoder complexity. Denoising requires O(1) operations per face removing a large portion of the computational cost. 
\color{black}

\begin{figure}
    \centering
    \includegraphics[width=\linewidth]{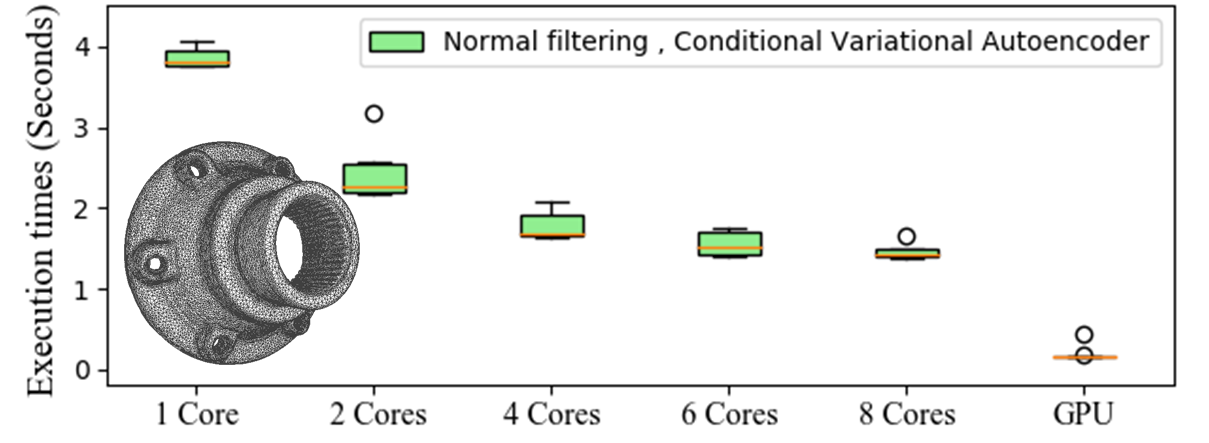}
    \caption{CVAE execution times for 100K faces model and different settings}
\label{fig:parallelization}
\end{figure}

\subsection{Impact of parallelization}

As subsection \ref{subsection:denoising} highlights, the input vector $\mathbf{Z}_i^{3(n+1) \times 1}$ contains the normal vector coordinates of $n$ neighbouring faces for a single patch. The matrix formulated for all patches can be expressed as $\mathbf{Z}^{n_f \times 3(n+1) }$ where $n_f$ is the number of faces for the processed model. Each patch $\mathbf{Z}_i$ is being processed separately through the same processing pipeline. Tensorflow already parallelizes this process and allows to control the number of used CPUs or GPUs. To further elaborate, a noisy model consisting of 100K faces, specifically the "carter" model, was denoised to measure the execution time for the autoencoder part. Six different settings were examined, namely, i) 1 CPU core, ii) 2 CPU cores, iii) 4 CPU cores, iv) 6 CPU cores, v) 8 CPU cores and vi) GPU only. For each different setting, 20 repetitions were performed. Figure \ref{fig:parallelization} presents boxplots summarizing the execution times distribution for each setting.

\subsection{Defect detection in an industrial setting}
\begin{figure}
\centering
\includegraphics[width=.95\linewidth]{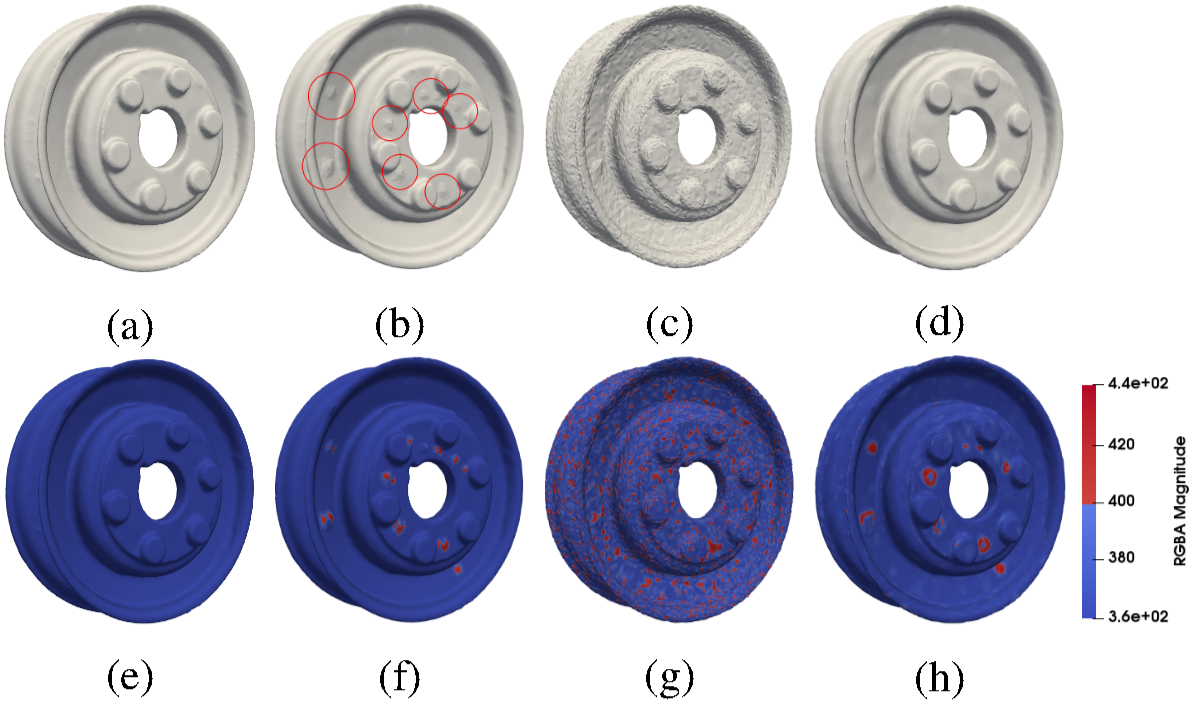}


\caption{\small Defect detection with saliency maps. a,e) Noise-free mesh b,f) Noise-free mesh mesh with surface defects c,g) Noisy mesh d,h) Denoised mesh}
\label{fig:saliency}
\end{figure}

Saliency maps \cite{lau2016tactile} are essential tools for reliable, accurate and computationally efficient 3D representations, by simplifying the representation of physical.
Figure \ref{fig:saliency} presents visual confirmation, that proper feature preserving denoising can facilitate defect detection in an industrial setting. 
The first row presents the 3D mesh and the second row, the result of the defect detection process. The color map is related to the Hausdorff distance of each mesh to the ground truth geometry presented in figure \ref{fig:saliency}.
The first column (a,e) presents the original object, the second column (b,f) presents the same object with surface defects that may have originated from the manufacturing process. As Figure \ref{fig:saliency} reveals, noise prohibits the detection of the defects. The third column (c,g) presents the same 3D mesh with defects and Gaussian noise that may have originated from scanning. Finally, the fourth column presents the denoised object, where the outcome of denoising facilitates accurate detection.

\section{Discussion}
In this work, we presented a fast data-driven denoising approach, applying conditional variational autoencoders to filter the normals of noisy 3D mesh patches. These patches are modelled by a scale, translation and rotation invariant patch descriptor exploited during the learning process. A computationally light post-processing step is employed only for fine-tuning purposes. Extensive evaluation studies verify the effectiveness of the proposed method, as compared to other robust and well-known state-of-the-art approaches. We summarize the benefits of the proposed approach in the following points.
\color{annotated}
\begin{inlinelistroman}
    \item The network can localize since, training and inference are performed in a sliding patch setup. The filtered face normal vector is generated by providing a local neighbouring region as input.
    \item It requires a relatively small training set since we employ a preprocessing approach that restricts the input data space.
    \item It is fully parallelizable.
    \item Evaluation studies indicate that it demonstrates lower complexity and execution times than other non-data-driven state-of-the-art approaches.
    \item It can be utilized in industrial applications for denoising of dense objects with features such as corners and edges.
    \item It is parameter-free since every used parameter is predefined and the user does not need to search for optimal values per model.
\end{inlinelistroman}
\color{black}
Furthermore, our study aims to contribute to the field of geometric deep learning where the sampling of the latent space is nonuniform, on contrary to image processing or voxelized versions of 3D volumes. The proposed approach removes the noise from noisy 3D meshes, given that the deep architecture is trained with the same noise distribution. This could prove extremely beneficial for the fast denoising of meshes generated from a particular scanning device operated at a manufacturing production line. However, different levels or types of noise require a different training process.

\color{annotated} We should also highlight that the size of the deep neural network has an impact on the denoising performance, in terms of execution times and energy efficiency. As a future step, model compression and acceleration strategies are yet to be investigated. They would allow a smaller network size and lower execution times.
\color{black}Further reduction of execution times could originate from the removal of post-processing steps. Specifically, patches are appropriately rotated and clustered. The bilateral filter copes with problems or irregularities, while extensive evaluation studies reveal that the required number of bilateral iterations is proportional to the noise level. Training the CVAE so that no post-processing is required could improve performance. Furthermore, employing robust clustering, to effectively group patches with similar distributions of normal coordinates, could efficiently reduce the size of training set or boost reconstruction accuracy.

\bibliographystyle{IEEEtran}
\bibliography{references}

\clearpage
\newpage

\onecolumn

\counterwithin*{equation}{subsection}
\section*{Supplementary Material}
\beginsupplement
\counterwithin*{equation}{section}

\subsection{Introduction}

\noindent This section provides supplementary material for the original paper, entitled as "Fast mesh denoising with data driven normal filtering using deep variational autoencoders". At the following sections, we present sections which are not included in the main manuscript. Specifically, the following analysis has been excluded from the revised manuscript due to space limitations, but we decide to provide it as supplementary material. However, If the reviewers believe that the specific analysis increases the quality of our manuscript, we could include it in a future version.

\subsection{Preliminaries}
\subsubsection{Deep autoencoders}

\noindent Deep autoencoders encompass a multi-layer neural network architecture where the hidden layers encode the input to a latent space \cite{vincent2008extracting} and decode the latter to a reconstructed input. A deep autoencoder is composed of two, symmetrical deep-belief networks \cite{hinton2009deep} that typically have three to five shallow layers for the encoding and the decoding part. The layers are restricted Boltzmann machines\cite{hinton2012practical}, the building blocks of deep-belief networks. For each hidden layer, input vector $\mathbf{x}_i \in \left[0,1\right]$ is mapped to a representation $\mathbf{y}_i \in \left[0,1\right]$ with a non-linear mapping,
\begin{equation}
    \mathbf{y}_i=f_\theta\left(\mathbf{x}_i\right)=s(\mathbf{W}_i\mathbf{x}_i+\mathbf{b}_i)
\end{equation}
The mapping is parameterized by $\theta_i=\{\mathbf{W}_i,\mathbf{b}_i\}$, where $\mathbf{W}_i$ is a 
weighting matrix and $\mathbf{b}_i$ is a bias vector. The resulting latent vector is mapped to a reconstructed output $\mathbf{z}_i\in \left[0,1\right]$.
\begin{equation}
    \mathbf{z}_i=g_\theta(\mathbf{y}_i)=s(\mathbf{W}'_i\mathbf{y}_i+\mathbf{b}'_i)
\end{equation}
where $\theta_i'=\{ \mathbf{W}_i',\mathbf{b}_i'\}$.
For each training epoch, defined as the full training pass over the entire dataset such that each example has been seen once, $\mathbf{x}_i$ is mapped to $\mathbf{y}_i$ and $\mathbf{y}_i$ to $\mathbf{z}_i$. The parameters are optimized to minimize the average reconstruction error. 
\begin{equation}
    \theta_i,\theta'_i={arg\,min}_{\theta_i,\theta'}\frac{1}{n}\sum_{i=1}^n L\left( \mathbf{x}_i , \mathbf{z}_i \right)
\end{equation}
where $L$ is a loss function and more commonly the L2-norm.
Likewise, the denoising autoencoder\cite{vincent2008extracting} aims to repair destroyed, corrupted or missing input by expressing $\mathbf{x}$ as $\mathbf{\tilde{x}}\sim q_D(\tilde{\mathbf{x}}|\mathbf{x})$ where $q_D(X)$ denotes the empirical distribution associated to $n$ training inputs. 
Consequently, $\mathbf{y}$ and $\mathbf{z}$ are defined as $\mathbf{y}=s(\mathbf{W}\tilde{\mathbf{x}}+\mathbf{b})$ and $\mathbf{z}=s(\mathbf{w}'\tilde{\mathbf{y}}+\mathbf{b}')$. The parameters are trained to minimize the average reconstruction cross entropy error perceiving $\mathbf{x}$ and $\mathbf{z}$ as vectors of probabilities  $\mathcal{L}_\mathbf{H}(\mathbf{x},\mathbf{z})=\mathbf{H}(\mathcal{B}_x,\mathcal{B}_z)$, where:
\begin{equation}
    \begin{split}
    \mathcal{L}_\mathbf{H}(\mathbf{x},\mathbf{z})=&\mathbf{H}(\mathcal{B}_x||\mathcal{B}_z) \\
    =&-\sum_{k=1}^d\left[\mathbf{x}_k\log \mathbf{z}_k +\left(1-\mathbf{x}_k \right) \log\left(1-\mathbf{z}_k\right)\right]
    \end{split}
\end{equation}

\subsection{Deep variational and deep conditional variational autoencoders}

\noindent Variational autoencoders (VAE) \cite{doersch2016tutorial,kingma2013auto} assume that the input vectors are generated by some random process of an unobserved continuous random variable $\mathbf{z}$. A $\mathbf{z}$ value is generated from some prior distribution $p_\theta(\mathbf{z})$ and a value $\mathbf{x}_i$ is generated from some conditional distribution $p_\theta(x)=\int p_\theta(z) p_\theta(x|z)dz$ which is intractable. Thus, a recognition model $q_\Phi(\mathbf{z}|\mathbf{x})$ can be employed so as to sample the possible values of $\mathbf{z}$ with a distribution from which the $\mathbf{z}$ could have been generated. $q_\Phi(\mathbf{z}|\mathbf{x})$ is referred to as a probabilistic encoder and $p_\theta(\mathbf{x}|\mathbf{z})$ as a probabilistic decoder. The parameters of the VAE are estimated efficiently by the stochastic gradient variational Bayes framework \cite{kingma2013auto} and the marginal likelihood is computed by the sum of marginal likelihoods of each point.
\begin{equation}
    \log p_\theta(\mathbf{x}_1,\cdots,\mathbf{x}_N)=\sum_{i=1}^N\log p_\theta(\mathbf{x}_i)
\end{equation}
and
\begin{equation}
    \log p_\theta(\mathbf{x}_i)=D_{KL}(q_\Phi(\mathbf{z}|\mathbf{x}_i)||p_\theta(\mathbf{z}|\mathbf{x}_i)))+\mathcal{L}(\theta,\Phi;\mathbf{X}_i)
\end{equation}
where $D_{KL}$ is the Kullback-Leibler divergence\cite{joyce2011kullback} and $\mathcal{L}(\theta,\Phi;\mathbf{X}_i)$ is the variational lower bound \cite{kingma2013auto}. The objective is to minimize:
\begin{equation}
\begin{split}
 &\log p_\theta(\mathbf{x}_i)-D_{KL}(q_\Phi(\mathbf{z}|\mathbf{x}_i)||p_\theta(\mathbf{z}|\mathbf{x}_i)))= \\ &\mathbb{E}[\log p(\mathbf{x}|z)]-D_{KL}(q_\Phi(\mathbf{z}|\mathbf{x}_i)||p_\theta(\mathbf{z})))
\end{split}
\end{equation}

An improvement of the VAE is the conditional variational autoencoder (CVAE) \cite{sohn2015learning} where the encoder and the decoder are conditioned under $\mathbf{x}$ and the label of $\mathbf{x}$ denoted as $c$.
 The CVAE objective is written as:
\begin{equation}
\begin{split}
 &\log p_\theta(\mathbf{x}_i|c_i)-D_{KL}(q_\Phi(\mathbf{z}|\mathbf{x}_i,c_i)||p_\theta(\mathbf{z}|\mathbf{x}_i)))= \\ &\mathbb{E}[\log p(\mathbf{x}_i|\mathbf{z}_i,c_i)]-D_{KL}(q_\Phi(\mathbf{z_i}|\mathbf{x}_i,c)||p_\theta(\mathbf{z}|c)))
\end{split}
\end{equation}

\end{document}